\documentclass[11pt,a4paper]{article}
\usepackage[hyperref]{naaclhlt2019}
\usepackage{times}
\usepackage{latexsym}

\usepackage{url}

\usepackage{xspace}
\usepackage{xargs} 
\usepackage[table]{colortbl}

\usepackage{booktabs}

\usepackage{array}
\usepackage{graphicx}
\usepackage{subfig}

\usepackage{amsmath}
\definecolor{OliveGreen}{cmyk}{0.64,0,0.95,0.40}
\usepackage{comment}

\usepackage{bbding}
\usepackage{pifont}
\usepackage{wasysym}
\usepackage{amssymb}
\usepackage{amsthm}
\usepackage{mathtools}

\usepackage{commath}
\usepackage{pdflscape}
\usepackage{tabularx,array}

\usepackage{enumerate}
\usepackage{enumitem}

\usepackage{algpseudocode}

\usepackage{times}
\usepackage{fancyhdr}
\usepackage{multirow}

\usepackage[
  separate-uncertainty = true,
  multi-part-units = repeat,
  detect-weight = true 
]{siunitx}
\usepackage{booktabs}
\usepackage{etoolbox,siunitx}
\robustify\bfseries
\usepackage{array,multirow,graphicx}

\usepackage{array}
\newcolumntype{P}[1]{>{\centering\arraybackslash}p{#1}}

\usepackage{xcolor}

\usepackage{setspace}

\usepackage[ruled,vlined]{algorithm2e}

\SetCommentSty{mycommfont}
\SetKwInput{KwInput}{Input}                
\SetKwInput{KwReturn}{return}

\usepackage[colorinlistoftodos,prependcaption,textsize=tiny]{todonotes}

\usepackage{marginnote}

\newcommandx{\siva}[2][1=]{\todo[linecolor=red,backgroundcolor=red!10,bordercolor=red,#1]{SR: #2}\xspace}
\newcommand{\ignore}[1]{}

\usepackage{ragged2e,array,booktabs}
\usepackage{bbding}
\usepackage{pifont}
\usepackage{wasysym}
\usepackage{amssymb}
\usepackage{balance}
\usepackage{pbox}

\aclfinalcopy 

\title{Question Generation for Evaluating Cross-Dataset Shifts in Multi-modal Grounding}

\author{Arjun R Akula \\
  University of California, Los Angeles \\
  {\tt aakula@ucla.edu}}

\date{}

\begin{document}
\maketitle
\begin{abstract}
    Visual question answering (VQA) is the multi-modal task of answering natural language questions about an input image. Through cross-dataset adaptation methods, it is possible to transfer knowledge from a source dataset with larger train samples to a target dataset where training set is limited.  Suppose a VQA model trained on one dataset train set fails in adapting to another, it is hard to identify the underlying cause of domain mismatch as there could exists a multitude of reasons such as image distribution mismatch and question distribution mismatch. At UCLA, we are working on a VQG module that facilitate in automatically generating OOD shifts that facilitates in systematically evaluating cross-dataset adaptation capabilities of VQA models.
\end{abstract}
\section{Background \& Motivation}
VQA is the challenging AI task of answering natural language questions about an input image~\cite{antol2015vqa,anderson2018bottom}. Several datasets have been proposed to measure the progress on this task such as VQA 2.0~\cite{goyal2017making}, VizWiz~\cite{gurari2018vizwiz}, Visual7w~\cite{zhu2016visual7w}, GQA~\cite{hudson2019gqa}, to name a few. Recently, self-attention and multi-modal pre-training based methods~\cite{tan2019lxmert,lu2019vilbert,cao-etal-2020-deformer} demonstrated superior performance on these datasets. Despite great progress, the state-of-the-art methods are found to be less effective when the distribution of $\langle image, question \rangle$ pairs in testing set are different from the training set~\cite{chao2018cross,akula2022discourse,gardner2020evaluating}, necessitating the importance of developing cross-dataset adaptation methods~\cite{DBLP:journals/corr/abs-1903-02252,carlson2003building,soricut2003sentence,lethanh2004generating}.

Through cross-dataset adaptation methods, it is possible to transfer knowledge from a source dataset with larger train samples to a target dataset where training set is limited. However, VQA datasets differ in the way they are collected, making them significantly different in the distribution of input visual and language features~\cite{chao2018cross}. These multi-modal distribution shifts make it difficult to measure adaptation capabilities and has yet to be well-studied. For example, consider a domain adaptation setting between VQA 2.0~\cite{goyal2017making} vs. VizWiz~\cite{gurari2018vizwiz} datasets.

Suppose a VQA model trained on VQA 2.0 train set fails in adapting (e.g. through fine-tuning) to Vizwiz. From this observation, it is difficult to identify the real cause of domain mismatch as there could exists a multitude of reasons such as (a) \textbf{image distribution} mismatch (e.g. \textit{VQA 2.0 consists of high quality images compared to Vizwiz}); (b) \textbf{question distribution} mismatch (e.g. \textit{VQA 2.0 questions are less conversational than Vizwiz questions}); (c) \textbf{insufficient sample size} (e.g. \textit{VQA 2.0 consists of relatively large number of training samples}); and (d) a combination of image and question distribution mismatches~\cite{DBLP:journals/corr/abs-1903-02252,akula20words,akula2019visual,akula2021crossvqa,akula2021measuring,akula2021robust,akula2020cocox,r2019natural,pulijala2013web,gupta2012novel}.

\begin{figure*}[t]
\centering
  \includegraphics[width=0.8\linewidth]{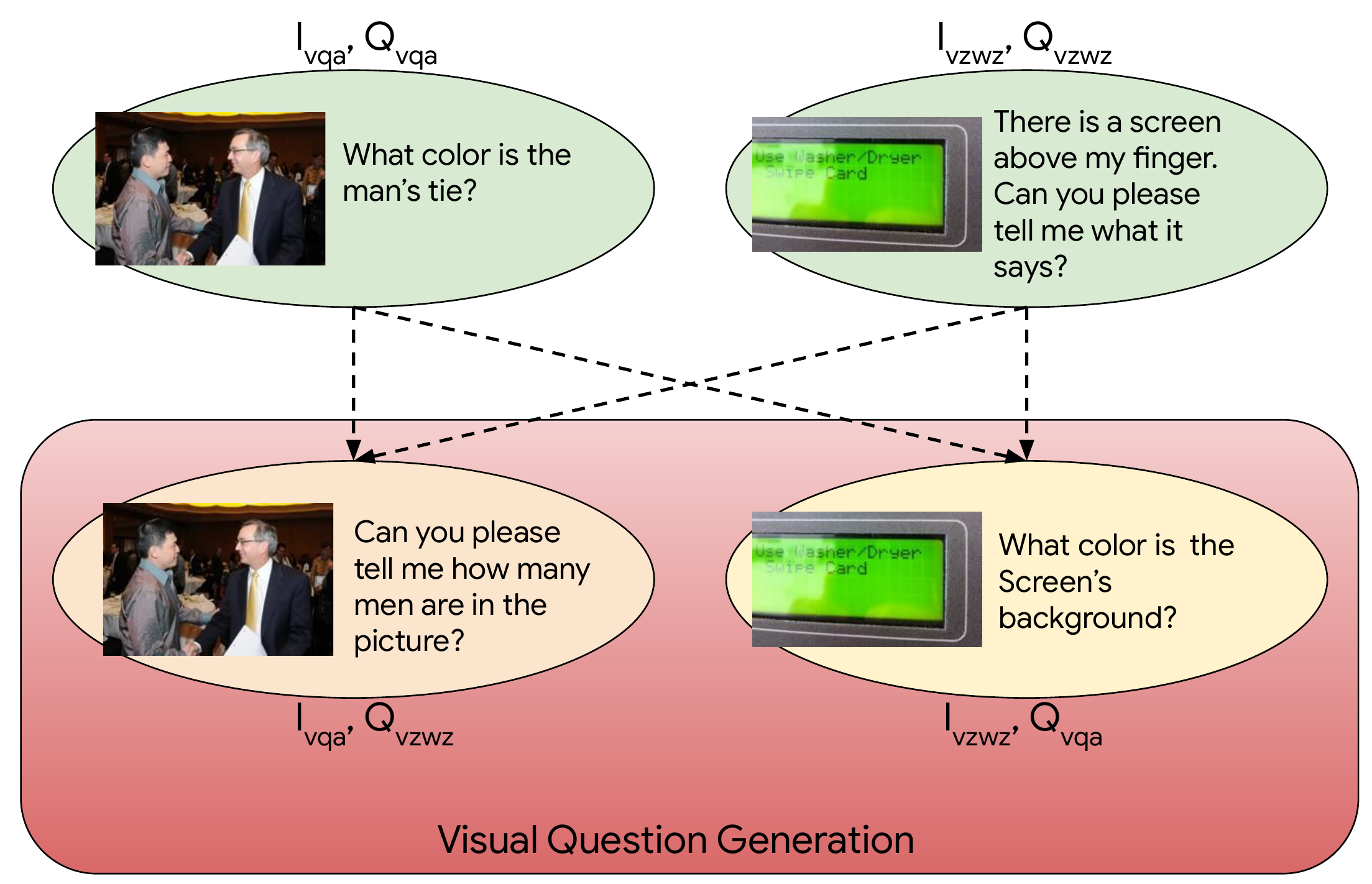}
  \caption{Examples from VQA 2.0 and VizWiz datasets. Existing works on domain adaptation compare the VQA model's performance on the test splits $\langle I_{vqa}, Q_{vqa} \rangle$ and $\langle I_{vzwz}, Q_{vzwz} \rangle$. In this work, we propose a Visual Question Generation module to generate additional tests sets $\langle I_{vqa}, Q_{vzwz} \rangle$ and $\langle I_{vzwz}, Q_{vqa} \rangle$ that facilitate in disentangled, incremental, and fine-grained evaluation of cross-dataset generalization.}~\label{fig:intro}
\end{figure*}

\section{Problem Definition}
There exists two methods to generate cross-datasets shifts:
\begin{enumerate}
\item \textbf{Human Annotations}: We can ask human annotators (eg. AMT turkers) to write VizWiz style queries for VQA 2.0 images and VQA 2.0 style questions for VizWiz images. Although, the quality of annotations will be high, this approach is costly and cannot scale to multiple datasets~\cite{akula2013novel,akula2018analyzing,akula2021mind,gupta2016desire,akula2019explainable,akula2021gaining,akula2019x,akula2020words}.
\item  \textbf{Visual Question Generation (VQG)}: In this work, instead of using human annotators, we propose a VQG module that facilitate in automatically generating OOD shifts for VQA datasets. This facilitates in systematically evaluating cross-dataset adaptation capabilities of VQA models. Specifically, using our VQG module, we generate additional test sets for source and target datasets by controlling and disentangling distribution shifts in vision and language features~\cite{akula2022effective,agarwal2018structured,akula2019natural,akula2015novel,palakurthi2015classification,agarwal2017automatic,dasgupta2014towards}. 

For example, as shown in Figure~\ref{fig:intro}, using $\langle image, question \rangle$ pairs from source and target test sets involve shift in both visual and language features such as $\langle I_{vqa}, Q_{vqa} \rangle$  to $\langle I_{vzwz}, Q_{vzwz} \rangle$. We augment these test sets to facilitate measuring adaptation capabilities of VQA models on incremental (systematic) shifts such as $\langle I_{vqa}, Q_{vqa} \rangle$ to $\langle I_{vqa}, Q_{vzwz} \rangle$ to $\langle I_{vzwz}, Q_{vzwz} \rangle$. 
\end{enumerate}

\section{Summary of Contributions}
Below we summarize our key contributions:
\begin{enumerate}
\item Proposing and Implementing a Visual Question Generation (VQG) module for generating questions and answers from the images.
\item Using our proposed VQG, we generate OOD test splits for VQA 2.0 and VizWiz datasets
\item We show that our generated OOD splits help in quantifying the systematic cross-dataset shifts in VQA models.
\end{enumerate}

\section{Approach}
We will leverage state-of-the-art implementations to train an end-to-end VQG model. Specifically, we use train sets of VQA 2.0 and VizWiz datasets and train VQG model mapping input image to questions with an additional dataset source indicator specifying the source of the sample. After we train our VQG, during inference, we change the dataset indicator to generate cross-dataset image and question pairs. For example, we pick VQA 2.0 image and provide the dataset indicator as VizWiz, for generating VizWiz style questions on VQA 2.0 images. We will experiment with several contextual features (such as adding bounding box annotations, pre-training on image captioning datasets, etc) to control the quality of the generated questions.

\section{Datasets and VQA Models}
We experiment with VQA 2.0 and VizWiz datasets. We use ViLBERT~\cite{lu2019vilbert}, a pretrain-then-transfer approach, as the state-of-the-art VQA model for our adaptation experiments.

\section{Initial Experiments}
We have first started selecting a state-of-the-art VQG model. We choose an existing implementation based on mutual information maximization~\cite{krishna2017visual}. We incorporated the following additional cues to the input to generate the cross-dataset splits: \\
(1) Source of the dataset (eg: VQA, VizWiz)\\
(2) First Three Words of Question: We found that VizWiz questions start with unique words such as ``Can you please" and ``Please tell me". So, we believe providing the first three words of the question as additional guidance would further help the model to understand the distribution style of questions that we like to generate.\\
(3) We are currently working on integrating answer categories as additional inputs to the module. The work by \cite{krishna2017visual} proposed16 categories on VQA dataset such as \textit{spatial}, \textit{binary} and \textit{count}. We are leveraging these categories to make the generated questions more diverse.

\textbf{Training VQA models on ViLBERT}: We have completed training VQA models using ViLBERT architecture. This step took us more time as this takes up to 5 days to train the model. Once we get decent questions generated using VQG, we will immediately start our adaptation experiments using the VQA models trained using ViLBERT.

\section{Conclusion}
We performed cross-dataset evaluation with VQA 2.0 and VizWiz datasets. To do this, we proposed a Visual Question Generation (VQG) module for generating questions and answers from the images.Our experiments demonstrate that our generated OOD splits help in quantifying the systematic cross-dataset shifts in VQA models.


\end{document}